# A NEW HUMANLIKE FACIAL ATTRACTIVENESS PREDICTOR WITH CASCADED FINE-TUNING DEEP LEARNING MODEL

*Jie Xu, Lianwen Jin\*, Lingyu Liang\*, Ziyong Feng, Duorui Xie*

South China University of Technology, Guangzhou 510641, China
\* Email: lianwen.jin@gmail.com; lianglysky@gmail.com

## ABSTRACT

This paper proposes a deep leaning method to address the challenging facial attractiveness prediction problem. The method constructs a convolutional neural network (CNN) for facial beauty prediction using a new deep cascaded fine tuning scheme with various face inputting channels, such as the original RGB face image, the detail layer image, and the lighting layer image. With a carefully designed CNN model of deep structure, large input size and small convolutional kernels, we have achieved a high prediction correlation of 0.88. This result convinces us that the problem of the facial attractiveness prediction can be solved by deep learning approach, and it also shows the important roles of the facial smoothness, lightness, and color information that involve in facial beauty evaluation, which is consistent with the result of recent psychology studies. Furthermore, we analyze the high-level features learnt by CNN through visualization of its hidden layers, and some interesting phenomena were observed. It is found that the contours and appearance of facial features (especially eyes and month) are the most significant facial attributes for facial attractiveness prediction, which is also consistent with visual perception intuition of human.

## 1. INTRODUCTION

Facial attractiveness has allured humans for centuries. Although previous studies [1-6] indicate that facial averageness, symmetry, and sexual dimorphism are significant factors that influence the facial attractiveness perception of a human, a universal definition of beauty remains elusive. Recently, automatic facial attractiveness prediction has led to ever-growing studies in computer vision and machine learning communities [7-14], which indicates that facial attractiveness prediction is an useful techniques for many applications, such as recommendation system, image retrieval and face beautification [21].

However, facial attractiveness prediction is still an unsolved problem. The challenge lies in two aspects. The first is due to the large appearance variance of a face that affects the beauty assessment, while the second is the lack of benchmark database for evaluation. Based on the latest proposed benchmark database SCUT-FBP [17], the goal of the paper is to find an effective method to construct an automatic facial attractiveness predictor that is consistent to the facial beauty perception of a human.

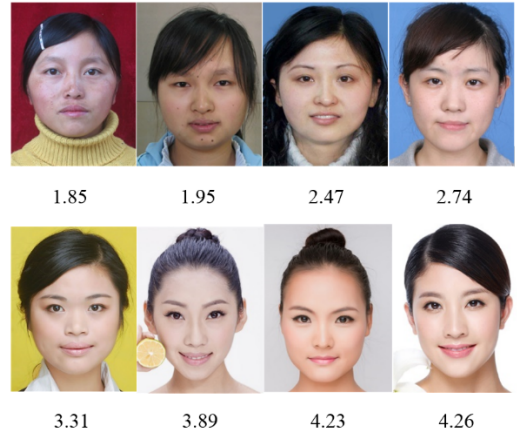

**Fig. 1.** Input faces and their facial beauty score predicted by our deep model. The score ranges between [1, 5], where the large score means the face is more attractive. The results indicate the deep facial beauty predictor is consistent to human perception.

According to our research, most previous studies address the problem using hand-designed feature with machine learning methods [7-14]. Kagian et al. [11] combined both geometric and appearance feature to build a predictor. Yan [12] proposed a cost-sensitive ordinal regression model for beauty assessment. Chiang et al. [13] extracted 3D facial features using a 3dMD scanner to train an adjusted fuzzy neural network, and high accuracy was achieved. Furthermore, dynamic features obtained from video clips were proved to be efficient with static facial features [14]. The previous studies indicate that the facial feature representation is essential for facial beauty prediction and analysis.

Recently, deep neural networks were proved to be an effective way to learn the high-level presentation for facial beauty prediction. Gan et al. [15] used deep self-taught learning with LBP or Gabor to extract the facial presentation, and the results indicate that SVM regression obtains better performance with the proposed feature. Wang et al. [16] used a pair of auto-encoders to obtain attractiveness-aware features and used a robust low-rank fusion framework to further enhance the performance.

The previous deep learning methods mostly consider the representation learning, while the predictor is still based on the shallow structure. In this paper, however, we perform the deep learning analysis for facial attractiveness prediction from a different perspective. We regard the high-level feature learning and the facial beauty predictor as a whole process. A cascaded fine tuning method with multiple facial feature is proposed to construct a CNN-based (convolutional neural network based) deep facial beauty predictor, which obtain the highest correlation of 0.88 in the SCUT-FBP database [17].

Experiments further convince us that deep learning with cascaded fine tuning is significant for facial attractiveness prediction, as shown in Fig. 1. Specifically, the deeper network with larger input images and smaller convolution kernels obtain better performance; the RGB color layer, lighting layer and detail layer image extracted from the original image provide effective visual cues for cascaded fine tuning of a deep model; the visualization of the hidden layers of the deep model indicates that the contours and appearance of facial features (especially eyes and month) are the significant facial attributes for facial attractiveness prediction.

The remainder of this paper is organized as follows. Section 2 describes the CNN models and propose the deep cascaded fine tuning facial attractiveness predictor. Section 3 presents and analyzes the experimental results and shows visualization images from our CNNs convolution layers. Section 4 concludes this paper by summarizing our findings.

## 2. CNN-BASED DEEP CASCADED FINE-TUNING FACIAL ATTRACTIVENESS PREDICTION

### 2.1. Architectures of the proposed CNN models

To explore the representation ability of a CNN for facial attractiveness prediction, we designed three basic convolutional neural networks, denoted as CNN-1, CNN-2, and CNN-3 respectively.

Specifically, all the networks are composed of one or more convolution layers, with each followed by a pooling layer. Fully connected layers are placed at the top of the layer. For facial attractiveness prediction, Euclidean distance is used in the loss function to evaluate the difference between the true labels and the predicted results. The dropout technique [22] is also used to avoid over-fitting. Input images are randomly cropped to a smaller size 10 times to generate samples during the training process, and a certain cropping ratio ensures facial integrity, to prevent information from being lost during the cropping process.

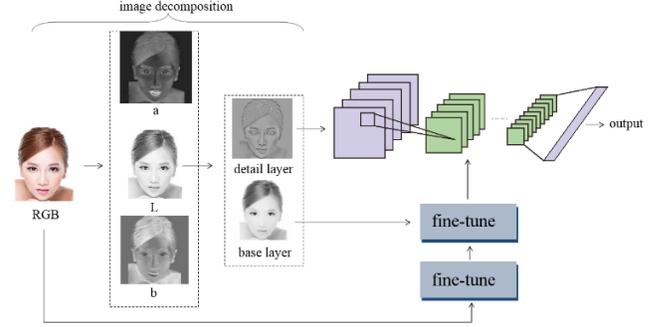

**Figure 2.** Framework of our facial attractiveness prodictor using deep cascaded fine tuning CNN model. First, we extract base layer and detail layer; second, we train a base model using the detail layer as input; third, we fine-tune the trained model with the RGB color channel and base layer channel.

The three CNNs are constructed with different depth of the net, input image sizes, and convolution kernel sizes, which are the key factors that determine their representation abilities. In our study, CNN-1 has the shallowest structure, while CNN-3 is the deepest. The detailed structural parameters are illustrated in Table 1. Note that all convolution layers are followed by a pooling layer with a factor of two, and the stride of each convolution layer is one.

### 2.2. CNN based deep cascaded fine tuning predictors with multiple facial channels

As described earlier, we believe the factors of smoothness, lightness, and color have a significant impact on facial attractiveness. Thus, we use image layer decomposition to extract facial features that correspond to the three factors, as shown in Figure 2.

First, we convert the input image into a CIELAB color space, in which the L channel contains lightness information while the a and b channels represent color information. Then, a weighted least squares filter is applied to the L channel to obtain a piecewise smooth base layer [23]. The detail layer is then defined as the difference between L and the smooth base layer. The base layer captures the larger scale variations in intensity, and the detail layer captures the smaller scale details in the image. Therefore, we obtain another three types of facial channels: the color channels of a and b, the base layer, and detail layer, which respectively represent the factors of color, lightness, and smoothness. Note that the precious RGB channels also represent color information, we compare their effectiveness later.

Facial attractiveness is not determined by just one factor. To efficiently use the extracted facial channels, we devised the deep cascaded fine tuning CNN model, which needed to be fine-tuned continuously using the multiple facial channels based on a trained model. The trained model can take any type of the facial channels as the input, but here we select the detail layer channel as demonstrated in Fig.2 for the reason that the detail channel performs the best for facial attractiveness prediction, according to Table 3.

**Table 1.** Detailed parameters of three basic CNNs

| Layers | CNN-1 Net | CNN-2 Net | CNN-3 Net |
|---|---|---|---|
| Input | 56x56, crop to 48x48 | 156x156, crop to 138x138 | 256x256, crop to 227x227 |
| Conv. Layer 1 | 50 maps 7x7 kernel | 50 maps 5x5 kernel | 50 maps 5x5 kernel |
| Conv. Layer 2 | 100 maps 6x6 kernel | 100 maps 5x5 kernel | 100 maps 5x5 kernel |
| Conv. Layer 3 | 150 maps 5x5 kernel | 150 maps 4x4 kernel | 150 maps 4x4 kernel |
| Conv. Layer 4 | - | 200 maps 4x4 kernel | 200 maps 4x4 kernel |
| Conv. Layer 5 | - | 250 maps 3x3 kernel | 250 maps 3x3 kernel |
| Conv. Layer 6 | - | - | 300 maps 2x2 kernel |
| Fully-connect 1 | 300 neurons | 300 neurons | 500 neurons |
| Fully-connect 2 | 1 neurons | 1 neurons | 1 neurons |

## 3. EXPERIMENTS AND ANALYSIS

Experiments were performed in a recently proposed benchmark database, named SCUT-FBP [17], which is carefully evaluated for facial beauty perception. The database contains 500 images of Asian female faces with the corresponding beauty ranking, ranging from 1 to 5. In every experiment, we randomly select 400 images from the database as the training data, and use the remaining 100 as the testing data. A Pearson correlation between the ground-truth average score rated by 75 persons and the predicted score of our CNN model is used to evaluate the performance of the model. The deep models in this paper are implemented on Caffe (Convolutional Architecture for Fast Feature Embedding), a deep learning framework developed by the Berkeley Vision and Learning Center [24].

### 3.1. Deep vs. Shallow Structures

To demonstrate the effectiveness of the deep model, we trained an end-to-end basic CNN model using the raw pixel of the image as input, with only 3 layers (Table 1, CNN-1).

We compare the basic CNN to other methods that are based on the traditional shallow structure, like SVM regression and Gaussian regression. The results are given in Table 2, which indicate that the deep basic CNN is superior to the other shallow machine learning methods.

**Table 2.** Comparison deep and shallow models

| Methods | SVM Regression[15] | Gaussian Regression[17] | CNN-1 |
|---|---|---|---|
| Corr. | 0.64 | 0.65 | 0.76 |

**Table 3.** Comparisons of the three basic CNNs

|  | 1 | 2 | 3 | 4 | 5 | Average |
|---|---|---|---|---|---|---|
| CNN-1 | 0.76 | 0.75 | 0.73 | 0.77 | 0.77 | 0.76 |
| CNN-2 | 0.81 | 0.81 | 0.79 | 0.78 | 0.83 | 0.80 |
| CNN-3 | 0.83 | 0.82 | 0.81 | 0.80 | 0.84 | **0.82** |

**Table 4.** Comparisons of different feature in basic model

| Feature | Corr. |
|---|---|
| a | 0.76 |
| b | 0.61 |
| RGB | 0.83 |
| Base layer | 0.79 |
| Detail layer | **0.85** |
| RGB + Base + Detail | 0.62 |

### 3.2. Comparison among different deep CNN structures

To find a proper structure of a deep model, we compare the three basic CNN models with different depth, input image sizes, and convolution kernel sizes, as shown in Table 1. Note that all the basic CNN only use the raw pixel of the image as input.

We performed a 5-folds cross validation to evaluate the three basic CNN respectively, as shown in Table2. The results indicate that the CNN-3 model, with the deepest structure, largest input size, and relative small convolutional kernel size at the top layer, obtain the best performance for facial beauty prediction. Therefore, we use the CNN-3 model as the basic network in the following experiments.

Recent psychology study indicates that the facial detail, color and lightness feature is significant for facial beauty perception. It inspires us to further improve the deep model using the facial feature instead of the raw pixel as network input. We decompose the input image into multiple layers that contains detail, color (RGB, a, b), and lightness feature

[23], and use these feature to train the deep model, as shown in Table 4.

The results show that not all the facial feature obtain the good performance as input for deep model, indicating that some facial factors may have more influence in the beauty prediction than other factors. Since the detail-layer-based deep beauty predictor obtain better performance than other feature, we use the detail layer as input for our basic thereafter deep model construction.

Furthermore, a phenomenon was observed that simple combination of multiple facial features did not produce good enough correlation for facial attractiveness prediction, which is against our expectations.

### 3.3. Evaluation for deep cascaded fine-tuning model

**Table 5.** Deep models with/without cascaded fine-tuning

| CNN Input | Firstly Fine-tuned | Secondly Fine-tuned | Corr. |
|---|---|---|---|
| Detail layer | Base layer | - | 0.82 |
| Detail layer | Base layer | RGB | 0.88 |

Facial beauty is effected by many factors, which means integrated more information into the detail facial layer may achieve better performance. We found that cascaded fine-tuning is a significant method to combine the information of the feature for facial attractiveness prediction.

We implement our CNN based deep cascaded fine tuning predictors by using the detail layer channel as input, and then fine-tuning it with the base layer and the RGB channels, as shown in Table 5. The results indicate that detail layer and the RGB and lightness layer are complementary in the deep fine-tuning model, which further improve the correlation to 0.88. The phenomena also make us to consider whether humans judge facial attractiveness in the same hierarchical manner, which merits further exploring.

### 3.4. Visualization of the CNN model

To better understand what the CNN has learned about facial attractiveness, we conducted visualizations on a trained CNN-3 model using CAFFE. Fig. 3 show some visualized images of the first four convolution layers and some interesting phenomena are observed. As demonstrated in Fig. 3, a CNN discards background information automatically and captures a faces contours and five sense organs, especially the eyes and mouth. Moreover, the first two convolution layers also retain some skin information. These factors learned by the CNN are highly relevant with what a human may consider when evaluating facial attractiveness, which establishes our work on the foundation of cognitive psychology and makes it grounded and meaningful.

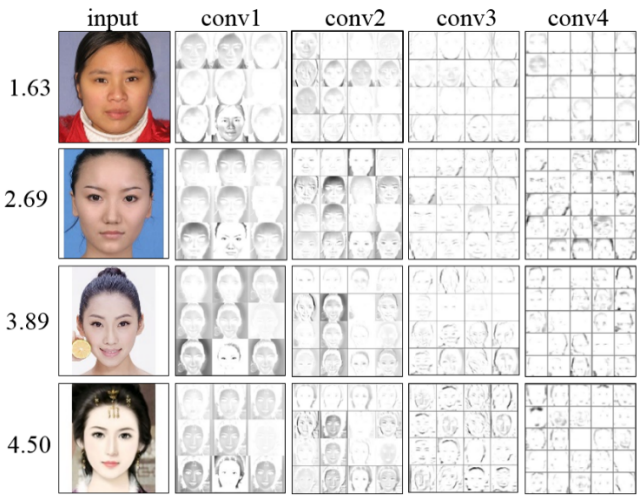

**Figure 3.** Visualization images of the first four convolution layers. The $5^{th}$ and $6^{th}$ convolution layers' visualization images are not interpretable, owing to their small sizes. The left digits are predicted scores using CNN-3, taking the pixels as inputs.

Furthermore, by comparing the visualizations of the four faces, we conclude that a more attractive face generates clearer visualization images, which means noise is discarded and useful messages are amplified by the CNN.

### 4. CONCLUTION

This paper addresses the facial attractiveness prediction problem using deep learning. A deep cascaded fine tuning predictor is proposed, which is based on convolutional neural network (CNN) using multiple face input channels. A very promising result of 88% correlation between our predictor outputs and the human scores is obtained, using the SCUTFBP database.

The analysis and experimental results indicate that for raw pixels as the inputs, the CNN model with deeper structure, larger input images and smaller convolution kernels, achieves better performance. Visualization shows that the hidden layers of the deep model can learn useful facial features that are consistent with visual perception of human, which may be useful for giving us clues to understand how the deep CNN model is used to handle the challenging artificial intelligence task such as facial attractiveness prediction.

There are some interesting questions still remain unsolved, such as why the simple combination of multiple facial features failed to produce reasonable good performance for facial attractiveness prediction, which need to be explored in future work.